\documentclass[11pt,a4paper]{article}
\usepackage[hyperref]{eacl2021}
\usepackage{times}
\usepackage{latexsym}

\usepackage[T1]{fontenc}
\usepackage{microtype}
\usepackage[T5]{fontenc}
\usepackage{graphicx}
\usepackage{amsmath}

\usepackage{multirow}
\aclfinalcopy 

\title{Trankit: A Light-Weight Transformer-based Toolkit for Multilingual Natural Language Processing}

\author{Minh Van Nguyen, Viet Lai, Amir Pouran Ben Veyseh, Thien Huu Nguyen\\
  Department of Computer and Information Science \\
  University of Oregon, Eugene, Oregon, USA \\
  \texttt{\{minhnv,vietl,apouranb,thien\}@cs.uoregon.edu}}

\date{}

\begin{document}
\maketitle
\begin{abstract}
We introduce \textbf{Trankit}, a light-weight \textbf{Tran}sformer-based Tool\textbf{kit} for multilingual Natural Language Processing (NLP). It provides a trainable pipeline for fundamental NLP tasks over 100 languages, and 90 pretrained pipelines for 56 languages. Built on a state-of-the-art pretrained language model, Trankit significantly outperforms prior multilingual NLP pipelines over sentence segmentation, part-of-speech tagging, morphological feature tagging, and dependency parsing while maintaining competitive performance for tokenization, multi-word token expansion, and lemmatization over 90 Universal Dependencies treebanks. Despite the use of a large pretrained transformer, our toolkit is still efficient in memory usage and speed. This is achieved by our novel plug-and-play mechanism with Adapters where a multilingual pretrained transformer is shared across pipelines for different languages. Our toolkit along with pretrained models and code are publicly available at: \url{https://github.com/nlp-uoregon/trankit}. A demo website for our toolkit is also available at: \url{http://nlp.uoregon.edu/trankit}. Finally, we create a demo video for Trankit at: \url{https://youtu.be/q0KGP3zGjGc}.
\end{abstract}

\section{Introduction}

Many efforts have been devoted to developing multilingual NLP systems to overcome language barriers \citep{aharoni-etal-2019-massively,liu-etal-2019-neural-cross,taghizadeh2020cross,zhu2020cross,kanayama-iwamoto-2020-universal,Nguyen:21improving}. A large portion of existing multilingual systems has focused on downstream NLP tasks that critically depend on upstream linguistic features, ranging from basic information such as token and sentence boundaries for raw text to more sophisticated structures such as part-of-speech tags, morphological features, and dependency trees of sentences (called fundamental NLP tasks). As such, building effective multilingual systems/pipelines for fundamental upstream NLP tasks to produce such information has the potentials to transform multilingual downstream systems.





There have been several NLP toolkits that concerns multilingualism for fundamental NLP tasks, featuring spaCy\footnote{\url{https://spacy.io/}}, UDify \citep{kondratyuk-straka-2019-75}, Flair \citep{akbik-etal-2019-flair}, CoreNLP \citep{manning-etal-2014-stanford}, UDPipe \citep{straka-2018-udpipe}, and Stanza \citep{qi-etal-2020-stanza}. However, these toolkits have their own limitations. spaCy is designed to focus on speed, thus it needs to sacrifice the performance. UDify and Flair cannot process raw text as they depend on external tokenizers. CoreNLP supports raw text, but it does not offer state-of-the-art performance. UDPipe and Stanza are the recent toolkits that leverage word embeddings, i.e., word2vec \citep{mikolov2013distributed} and fastText \citep{bojanowski-etal-2017-enriching}, to deliver current state-of-the-art performance for many languages. However, Stanza and UDPipe's pipelines for different languages are trained separately and do not share any component, especially the embedding layers that account for most of the model size. This makes their memory usage grow aggressively as pipelines for more languages are simultaneously needed and loaded into the memory (e.g., for language learning apps). Most importantly, none of such toolkits have explored contextualized embeddings from pretrained transformer-based language models that have the potentials to significantly improve the performance of the NLP tasks, as demonstrated in many prior works \citep{devlin-etal-2019-bert,liu2019roberta,conneau-etal-2020-unsupervised}.

\begin{figure*}
    \centering
    \includegraphics[scale=0.7]{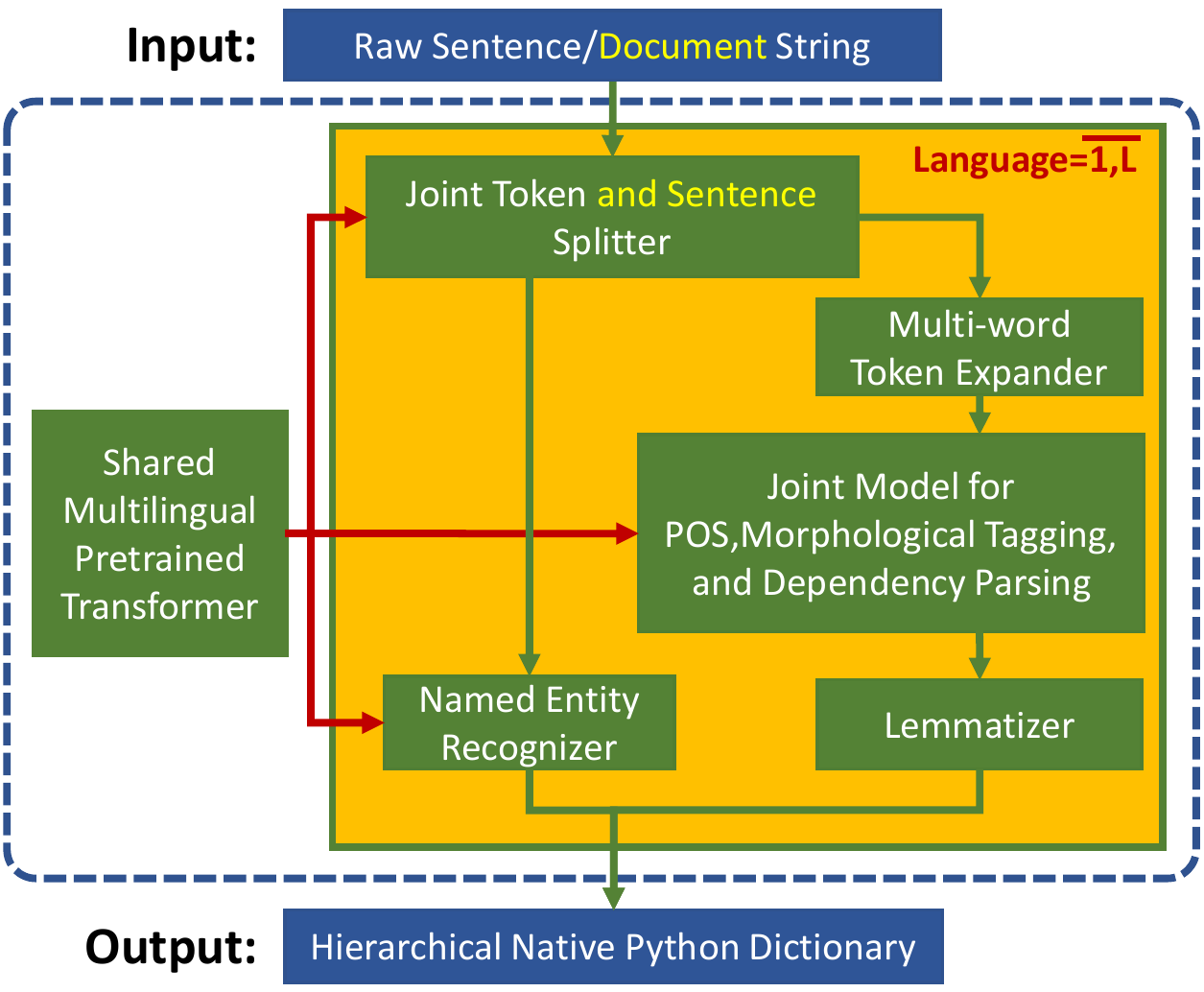}
    \caption{Overall architecture of Trankit. A single multilingual pretrained transformer is shared across three components (pointed by the red arrows) of the pipeline for different languages.}
    \label{fig:architecture}
\end{figure*}

In this paper, we introduce \textbf{Trankit}, a multilingual Transformer-based NLP Toolkit that overcomes such limitations. Our toolkit can process raw text for fundamental NLP tasks, supporting $56$ languages with $90$ pre-trained pipelines on $90$ treebanks of the Universal Dependency v2.5 \citep{11234/1-3105}. By utilizing the state-of-the-art multilingual pretrained transformer $\textrm{\textit{XLM-Roberta}}$ \citep{conneau-etal-2020-unsupervised}, Trankit advances state-of-the-art performance for sentence segmentation, part-of-speech (POS) tagging, morphological feature tagging, and dependency parsing while achieving competitive or better performance for tokenization, multi-word token expansion, and lemmatization over the 90 treebanks. It also obtains competitive or better performance for named entity recognition (NER) on 11 public datasets.

Unlike previous work, {\it our token and sentence splitter is wordpiece-based} instead of character-based to better exploit contextual information, which are beneficial in many languages. Considering the following sentence:


\begin{quote}
    \small \textit{``John Donovan from \underline{Argghhh!} has put out a excellent slide show on what was actually found and fought for in Fallujah.''}
\end{quote}
As such, Trankit correctly recognizes this as a single sentence while character-based sentence splitters of Stanza and UDPipe are easily fooled by the exclamation mark \textit{``!''}, treating it as two separate sentences. To our knowledge, this is the first work to successfully build a wordpiece-based token and sentence splitter that works well for 56 languages.


Figure \ref{fig:architecture} presents the overall architecture of Trankit pipeline that features three novel transformer-based components for: (i) the joint token and sentence splitter, (ii) the joint model for POS tagging, morphological tagging, dependency parsing, and (iii) the named entity recognizer. One potential concern for our use of a large pretrained transformer model (i.e., {\it XML-Roberta}) in Trankit involves GPU memory where different transformer-based components in the pipeline for one or multiple languages must be simultaneously loaded into the memory to serve multilingual tasks. This could extensively consume the memory if different versions of the large pre-trained transformer (finetuned for each component) are employed in the pipeline. As such, we introduce a novel plug-and-play mechanism with Adapters to address this memory issue. Adapters are small networks injected inside all layers of the pretrained transformer model that have shown their effectiveness as a light-weight alternative for the traditional finetuning of pretrained transformers \citep{houlsby2019parameter,peters-etal-2019-tune,pfeiffer-etal-2020-adapterhub,pfeiffer-etal-2020-mad}. In Trankit, a set of adapters (for transfomer layers) and task-specific weights (for final predictions) are created for each transformer-based component for each language while only one single large multilingual pretrained transformer is shared across components and languages. Adapters allow us to learn language-specific features for tasks. During training, the shared pretrained transformer is fixed while only the adapters and task-specific weights are updated. At inference time, depending on the language of the input text and the current active component, the corresponding trained adapter and task-specific weights are activated and plugged into the pipeline to process the input. This mechanism not only solves the memory problem but also substantially reduces the training time.

\section{Related Work}
There have been works using pre-trained transformers to build models for character-based word segmentation for Chinese \citep{yang2019bert,tian-etal-2020-joint-chinese,che2020n}; POS tagging for Dutch, English, Chinese, and Vietnamese \citep{de2019bertje,tenney-etal-2019-bert,tian-etal-2020-joint-chinese,che2020n,nguyen-tuan-nguyen-2020-phobert}; morphological feature tagging for Estonian and Persian \citep{kittask2020evaluating,mohseni-tebbifakhr-2019-morphobert}; and dependency parsing for English and Chinese \citep{tenney-etal-2019-bert,che2020n}. However, all of these works are only developed for some specific language, thus potentially unable to support and scale to the multilingual setting.

Some works have designed multilingual transformer-based systems via multilingual training on the combined data of different languages \citep{tsai-etal-2019-small,kondratyuk-straka-2019-75,ustun-etal-2020-udapter}. However, multilingual training is suboptimal (see Section \ref{sec:exp}). Also, these systems still rely on external resources to perform tokenization and sentence segmentation, thus unable to consume raw text. To our knowedge, this is the first work to successfully build a multilingual transformer-based NLP toolkit where different transformer-based models for many languages can be simultaneously loaded into GPU memory and process raw text inputs of different languages.

\section{Design and Architecture}


\begin{figure}[ht]
    \centering
    \includegraphics[scale=0.6]{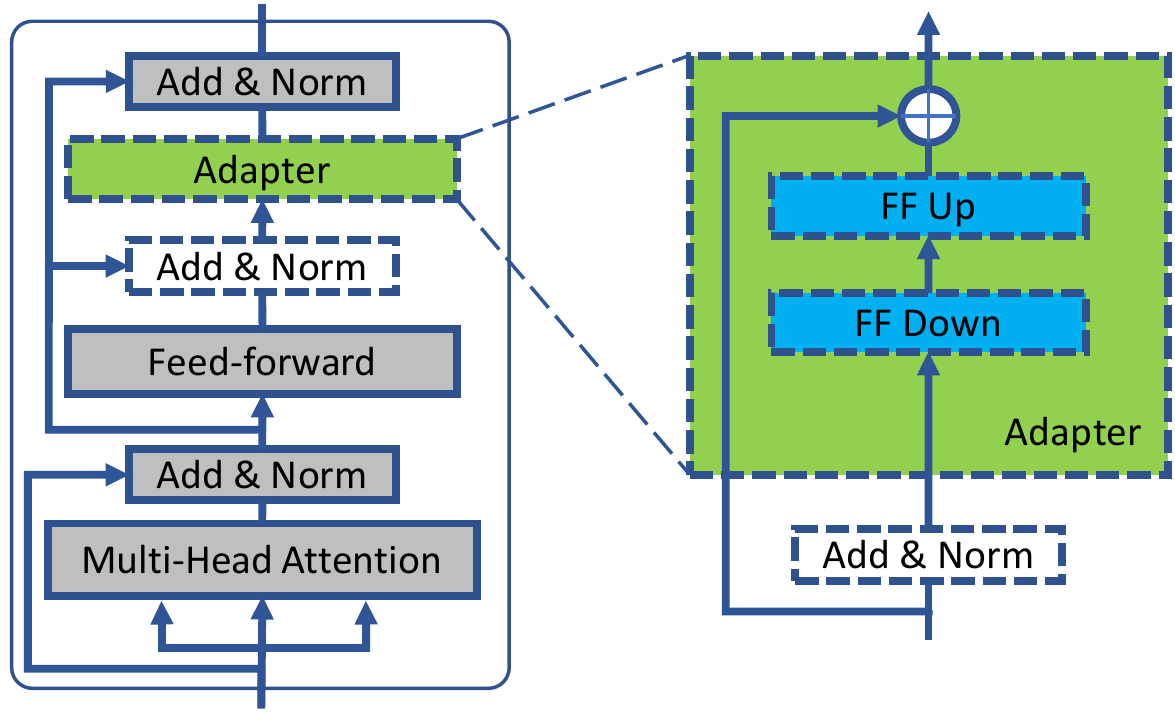}
    \caption{\textbf{Left}: location of an adapter (green box) inside a layer of the pretrained transformer.  Gray boxes represent the original components of a transformer layer. \textbf{Right}: the network architecture of an adapter.}
    \label{fig:adapters}
\end{figure}

\noindent \textbf{Adapters.} Adapters play a critical role in making Trankit memory- and time-efficient for training and inference. Figure \ref{fig:adapters} shows the architecture and the location of an adapter inside a layer of transformer. We use the adapter architecture proposed by \citep{pfeiffer-etal-2020-adapterhub,pfeiffer-etal-2020-mad}, which consists of two projection layers \textrm{Up} and \textrm{Down} (feed-forward networks), and a residual connection.
\begin{equation}
\small
    c_i = \textrm{AddNorm}(r_i),  h_i = \textrm{Up}(\textrm{ReLU}(\textrm{Down}(c_i))) + r_i
\end{equation}
where $r_i$ is the input vector from the transformer layer for the adapter and $h_i$ is the output vector for the transformer layer $i$. During training, all the weights of the pretrained transformer (i.e., gray boxes) are fixed and only the adapter weights of two projection layers and the task-specific weights outside the transformer (for final predictions) are updated. As demonstrated in Figure \ref{fig:architecture}, Trankit involves six components described as follows.



\vspace{0.3cm}

\noindent \textbf{Multilingual Encoder with Adapters.} This is our core component that is shared across different transformer-based components for different languages of the system. Given an input raw text $s$, we first split it into substrings by spaces. Afterward, Sentence Piece, a multilingual subword tokenizer \citep{kudo-richardson-2018-sentencepiece,kudo-2018-subword}, is used to further split each substring into wordpieces. By concatenating wordpiece sequences for substrings, we obtain an overall sequence of wordpieces $\textbf{w}= [w_1, w_2, \ldots, w_K]$ for $s$. In the next step, $\textbf{w}$ is fed into the pretrained transformer, which is already integrated with adapters, to obtain the wordpiece representations:
\begin{equation} \label{eq:encoder}
\small
    x^{l,m}_{1:K} = \textrm{Transformer}(w_{1:K}; \theta_{AD}^{l,m})
\end{equation}
Here, $\theta_{AD}^{l,m}$ represents the adapter weights for language $l$ and component $m$ of the system. As such, we have specific adapters in all transformer layers for each component $m$ and language $l$. Note that if $K$ is larger than the maximum input length of the pretrained transformer (i.e., 512), we further divide \textbf{w} into consecutive chunks; each has the length less than or equal to the maximum length. The pretrained transformer is then applied over each chunk to obtain a representation vector for each wordpiece in $\textbf{w}$. Finally, $x^{l,m}_{1:K}$ will be sent to component $m$ to perform the corresponding task.


\vspace{0.3cm}

\noindent \textbf{Joint Token and Sentence Splitter.} Given the wordpiece representations $x^{l,m}_{1:K}$ for this component, each vector $x^{l,m}_i$ for $w_i \in \textbf{w}$ will be consumed by a feed-forward network with softmax in the end to predict if $w_i$ is the end of a single-word token, the end of a multi-word token, or the end of a sentence. The predictions for all wordpieces in $\textbf{w}$ will then be aggregated to determine token, multi-word token, and sentence boundaries for $s$.



\vspace{0.3cm}

\noindent \textbf{Multi-word Token Expander.} This component is responsible for expanding each detected multi-word token (MWT) into multiple syntactic words\footnote{For languages (e.g., English, Chinese) that do not require MWT expansion, tokens and words are the same concepts.}. We follow Stanza to deploy a character-based seq2seq model for this component. This decision is made based on our observation that the task is done best at character level, and the character-based model (with character embeddings) is very small.

\begin{table*}[ht]
\centering
\scalebox{0.85}{
\begin{tabular}{l|l|c|c|c|c|c|c|c|c|c}
\multicolumn{1}{c|}{Treebank}          & \multicolumn{1}{c|}{System} & Tokens         & Sents.         & Words          & UPOS           & XPOS           & UFeats         & Lemmas         & UAS            & LAS            \\ \hline
\multirow{2}{*}{Overall (90 treebanks)} & {\bf Trankit}                     & 99.23          & \textbf{91.82} & \textbf{99.02} & \textbf{95.65} & \textbf{94.05} & \textbf{93.21} & \textbf{94.27} & \textbf{87.06} & \textbf{83.69} \\ \cline{2-11}
                                        & Stanza                      & \textbf{99.26} & 88.58          & 98.90          & 94.21          & 92.50          & 91.75          & 94.15          & 83.06          & 78.68          \\ \hline
\multirow{3}{*}{Arabic-PADT}            & {\bf Trankit}                     & 99.93          & \textbf{96.59} & \textbf{99.22} & \textbf{96.31} & \textbf{94.08} & \textbf{94.28} & \textbf{94.65} & \textbf{88.39} & \textbf{84.68} \\ \cline{2-11} 
                                        & Stanza                      & \textbf{99.98} & 80.43          & 97.88          & 94.89          & 91.75          & 91.86          & 93.27          & 83.27          & 79.33          \\ \cline{2-11} 
                                        & UDPipe                      & 99.98          & 82.09          & 94.58          & 90.36          & 84.00          & 84.16          & 88.46          & 72.67          & 68.14          \\ \hline
\multirow{3}{*}{Chinese-GSD}            & {\bf Trankit}                     & \textbf{97.01} & \textbf{99.7}  & \textbf{97.01} & \textbf{94.21} & \textbf{94.02} & \textbf{96.59} & \textbf{97.01} & \textbf{85.19} & \textbf{82.54} \\ \cline{2-11} 
                                        & Stanza                      & 92.83          & 98.80          & 92.83          & 89.12          & 88.93          & 92.11          & 92.83          & 72.88          & 69.82          \\ \cline{2-11} 
                                        & UDPipe                      & 90.27          & 99.10          & 90.27          & 84.13          & 84.04          & 89.05          & 90.26          & 61.60          & 57.81          \\ \hline
\multirow{4}{*}{English-EWT}            & {\bf Trankit}                     & 98.48          & \textbf{88.35} & 98.48          & \textbf{95.95} & \textbf{95.71} & \textbf{96.26} & 96.84          & \textbf{90.14} & \textbf{87.96} \\ \cline{2-11} 
                                        & Stanza                      & \textbf{99.01} & 81.13          & \textbf{99.01} & 95.40          & 95.12          & 96.11          & \textbf{97.21} & 86.22          & 83.59          \\ \cline{2-11} 
                                        & UDPipe                      & 98.90          & 77.40          & 98.90          & 93.26          & 92.75          & 94.23          & 95.45          & 80.22          & 77.03          \\ \cline{2-11} 
                                        & spaCy   & 97.44                               & 63.16                               & 97.44                               & 86.99                               & 91.05                               & -                                   & 87.16                               & 55.38                               & 37.03                               \\ \hline
\multirow{4}{*}{French-GSD}             & {\bf Trankit}                     & \textbf{99.7}  & \textbf{96.63} & \textbf{99.66} & \textbf{97.85} & -              & \textbf{97.16} & \textbf{97.80} & \textbf{94.00} & \textbf{92.34} \\ \cline{2-11} 
                                        & Stanza                      & 99.68          & 94.92          & 99.48          & 97.30          & -              & 96.72          & 97.64          & 91.38          & 89.05          \\ \cline{2-11} 
                                        & UDPipe                      & 99.68          & 93.59          & 98.81          & 95.85          & -              & 95.55          & 96.61          & 87.14          & 84.26          \\ \cline{2-11} 
                                        & spaCy   & 99.02                               & 89.73                               & 94.81                               & 89.67                               & -                                   & -                                   & 88.55                               & 75.22                               & 66.93                               \\ \hline
\multirow{4}{*}{Spanish-Ancora}         & {\bf Trankit}                     & 99.94          & \textbf{99.13} & 99.93          & \textbf{99.02} & \textbf{98.94} & \textbf{98.8}  & \textbf{99.17} & \textbf{94.11} & \textbf{92.41} \\ \cline{2-11} 
                                        & Stanza                      & \textbf{99.98} & 99.07          & \textbf{99.98} & 98.78          & 98.67          & 98.59          & 99.19          & 92.21          & 90.01          \\ \cline{2-11} 
                                        & UDPipe                      & 99.97          & 98.32          & 99.95          & 98.32          & 98.13          & 98.13          & 98.48          & 88.22          & 85.10          \\ \cline{2-11} 
                                        & spaCy                       & 99.95          & 97.54          & 99.43          & 93.43          & -              & -              & 80.02          & 89.35          & 83.81          \\ 
\end{tabular}}

\caption{Systems' performance on test sets of the Universal Dependencies v2.5 treebanks. Performance for Stanza, UDPipe, and spaCy is obtained using their public pretrained models. The overall performance for Trankit and Stanza is computed as the macro-averaged F1 over 90 treebanks. Detailed performance of Trankit for 90 supported treebanks can be found at \href{https://trankit.readthedocs.io/en/latest/performance.html}{our documentation page}.}
\label{tab:ud-performance}
\end{table*}




\vspace{0.3cm}

\noindent \textbf{Joint Model for POS Tagging, Morphological Tagging and Dependency Parsing.} In Trankit, given the detected sentences and tokens/words, we use a single model to jointly perform POS tagging, morphological feature tagging and dependency parsing at sentence level. Joint modeling mitigates error propagation, saves the memory, and speedups the system. In particular, given a sentence, the representation for each word is computed as the average of its wordpieces' transformer-based representations in $x^{l,m}_{1:K}$. Let $t_{1:N}=[t_1, t_2, \ldots, t_N]$ be the representations of the words in the sentence. We compute the following vectors using feed-forward networks $\textrm{FFN}_*$:
\begin{equation*}
\small 
    \begin{split}
        r^{upos}_{1:N} &= \textrm{FFN}_{upos}(t_{1:N}), r^{xpos}_{1:N} = \textrm{FFN}_{xpos}(t_{1:N})\\
        r^{ufeats}_{1:N} &= \textrm{FFN}_{ufeats}(t_{1:N}), r^{dep}_{0:N} = [x_{cls}; \textrm{FFN}_{dep}(t_{1:N})]
    \end{split}
\end{equation*}
Vectors for the words in $r^{upos}_{1:N}$, $r^{xpos}_{1:N}$, $r^{ufeats}_{1:N}$ are then passed to a softmax layer to make predictions for UPOS, XPOS, and UFeats tags for each word. For dependency parsing, we use the classification token \texttt{<s>} to represent the root node, and apply Deep Biaffine Attention \citep{dozat2016deep} and the Chu-Liu/Edmonds algorithm \citep{chu1965shortest,edmonds1967optimum} to assign a syntactic head and the associated dependency relation to each word in the sentence.

\vspace{0.3cm}

\noindent \textbf{Lemmatizer.} This component receives sentences and their predicted UPOS tags to produce the canonical form for each word. We also employ a character-based seq2seq model for this component as in Stanza.

\vspace{0.3cm}

\noindent \textbf{Named Entity Recognizer.} Given a sentence, the named entity recognizer determines spans of entity names by assigning a BIOES tag to each token in the sentence. We deploy a standard sequence labeling architecture using transformer-based representations for tokens, involving a feed-forward network followed by a Conditional Random Field.

\section{Usage}

Detailed documentation for Trankit can be found at: \url{https://trankit.readthedocs.io}.

\vspace{0.3cm}

\noindent \textbf{Trankit Installation}. Trankit is written in Python and available on PyPI: \url{https://pypi.org/project/trankit/}. Users can install our toolkit via pip using:

\begin{quote}
    \centering
    \texttt{pip install trankit}
\end{quote}

\noindent \textbf{Initialize a Pipeline}. Lines 1-4 in Figure \ref{fig:multilingual} shows how to initialize a pretrained pipeline for English; it is instructed to run on GPU and store downloaded pretrained models to the specified cache directory. Trankit will not download pretrained models if they already exist.






\vspace{0.3cm}

\noindent \textbf{Multilingual Usage.} Figure \ref{fig:multilingual} shows how to initialize a multilingual pipeline and process inputs of different languages in Trankit:
\begin{figure}[ht]
    \centering
    \includegraphics[scale=0.33]{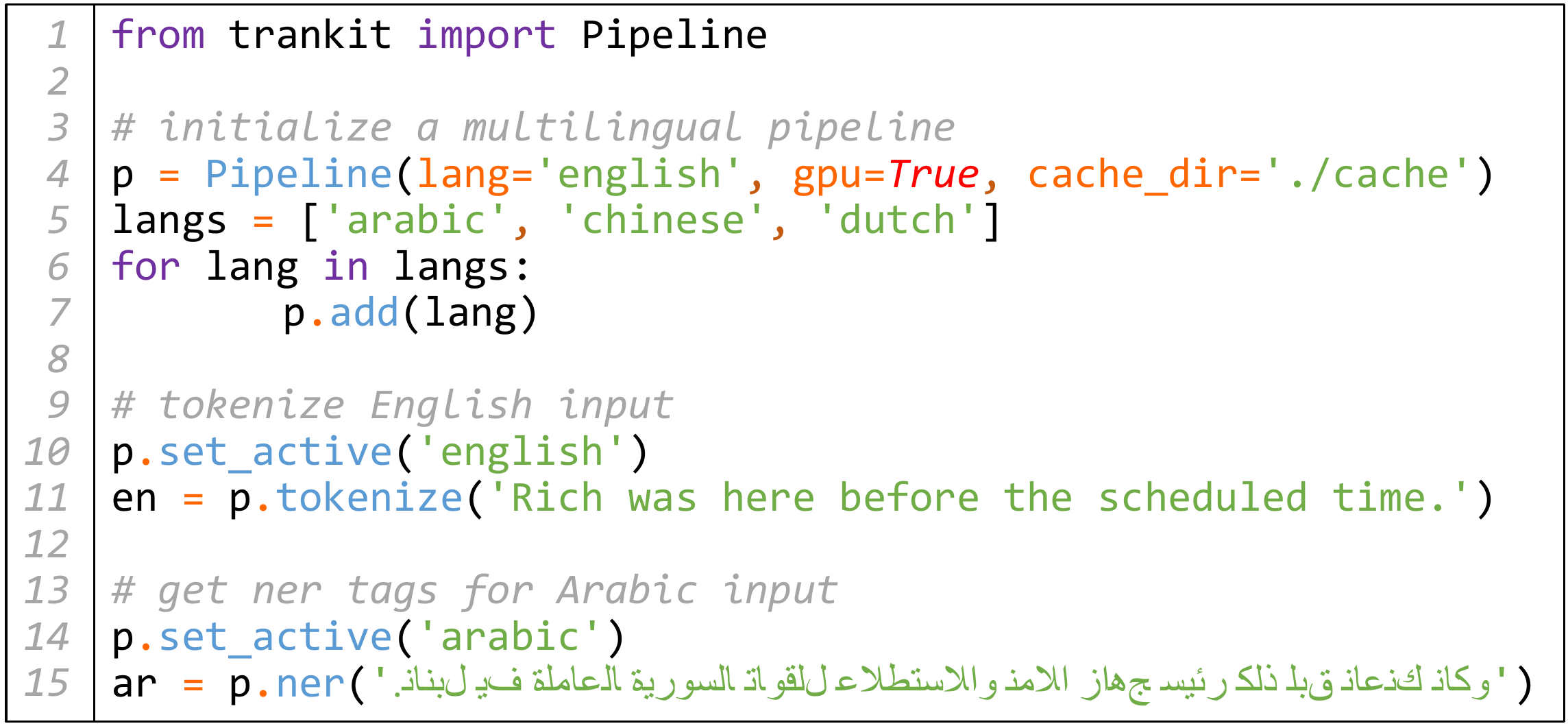}
    \caption{Multilingual pipeline initialization.}
    \label{fig:multilingual}
\end{figure}



\vspace{0.3cm}

\begin{figure}[ht]
    \centering
    \includegraphics[scale=0.33]{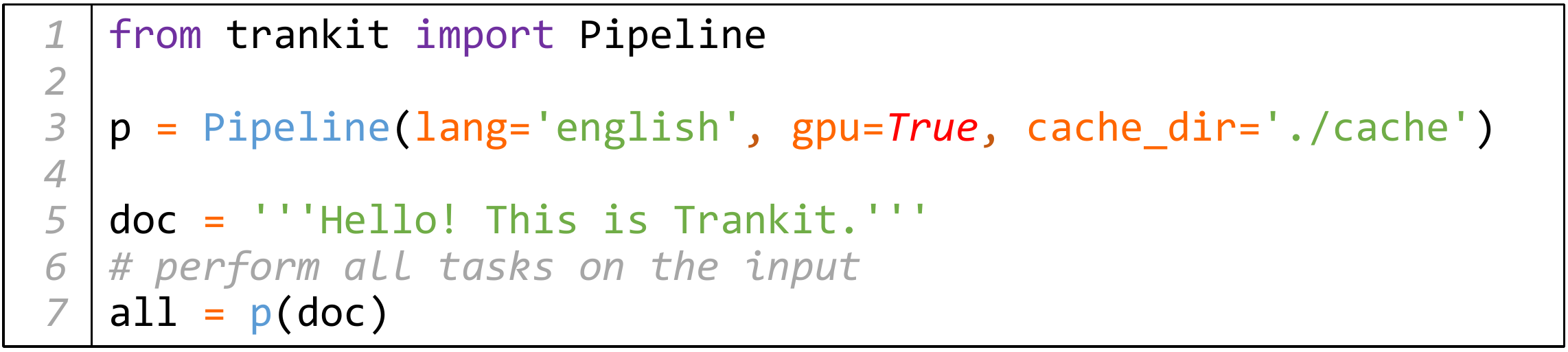}
    \caption{A function performing all tasks on the input.}
    \label{fig:basic}
\end{figure}

\begin{figure}[ht]
    \centering
    \includegraphics[scale=0.34]{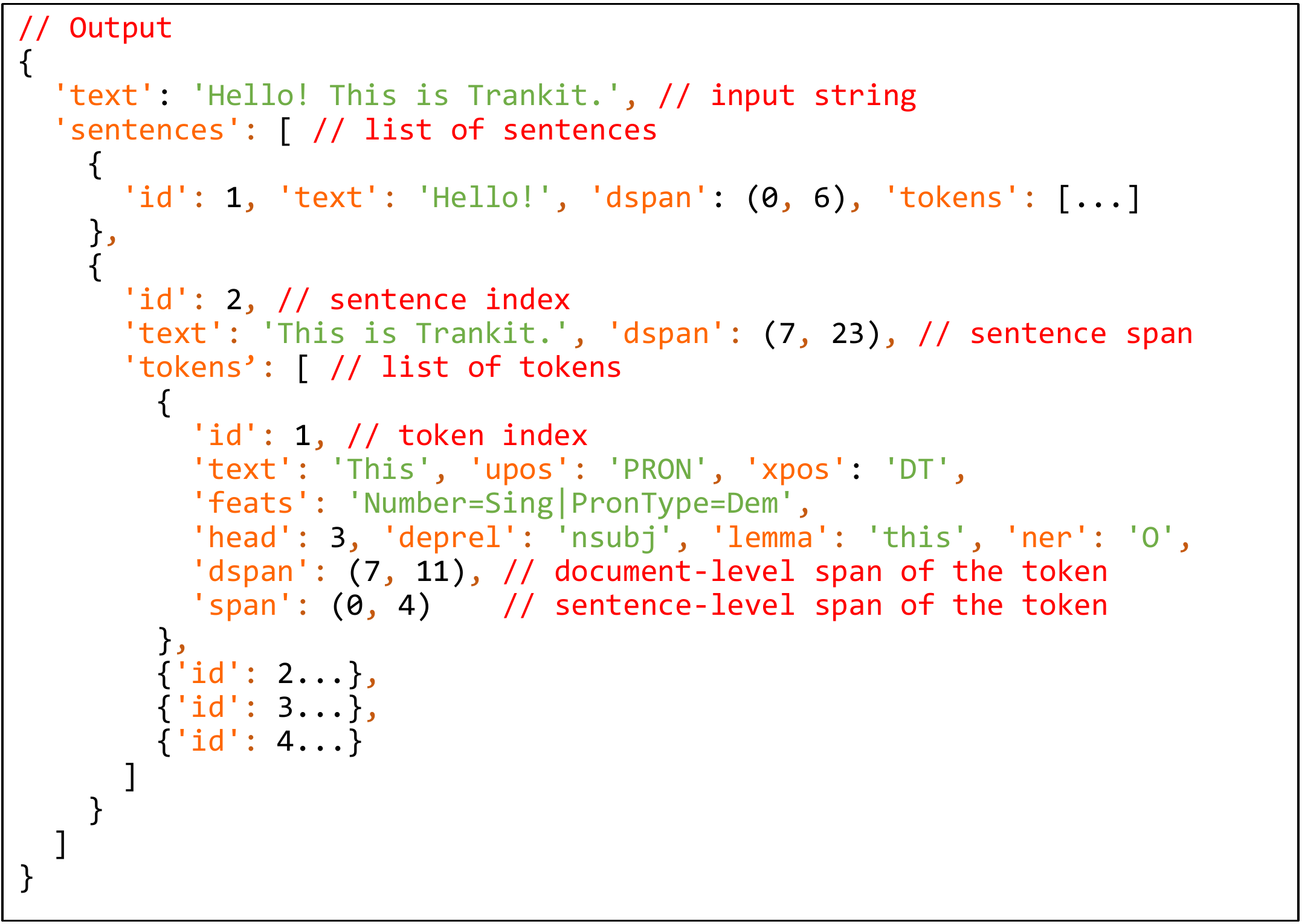}
    \caption{Output from Trankit. Some parts are collapsed to improve visualization.}
    \label{fig:output}
\end{figure}

\noindent \textbf{Basic Functions.} Trankit can process inputs which are untokenized (raw) or pretokenized strings, at both sentence and document levels. Figure \ref{fig:basic} illustrates a simple code to perform all the supported tasks for an input text. We organize Trankit's outputs into hierarchical native Python dictionaries, which can be easily inspected by users. Figure \ref{fig:output} demonstrates the outputs of the command line $6$ in Figure \ref{fig:basic}.


\vspace{0.3cm}

\begin{figure}[ht]
    \centering
    \includegraphics[scale=0.33]{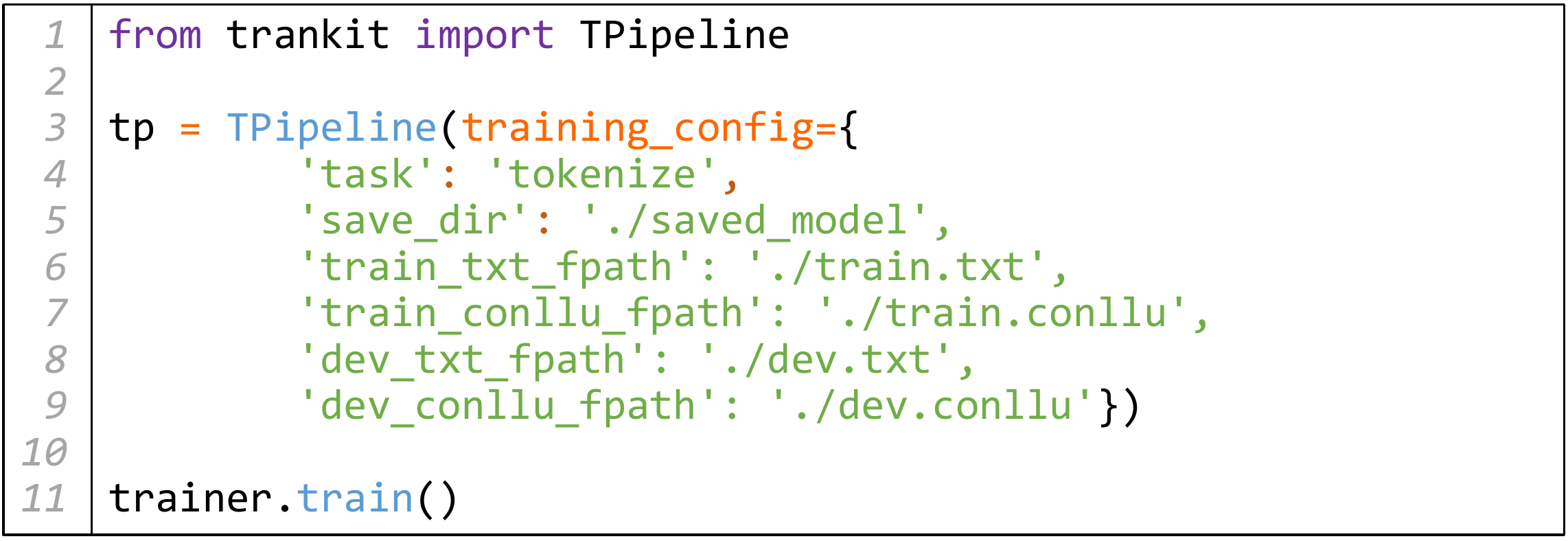}
    \caption{Training a token and sentence splitter using the CONLL-U formatted data \cite{nivre-etal-2020-universal}.}
    \label{fig:train}
\end{figure}

\noindent \textbf{Training your own Pipelines}. Trankit also provides a trainable pipeline for 100 languages via the class \texttt{TPipeline}. This ability is inherited from the XLM-Roberta encoder which is pretrained on those languages. Figure \ref{fig:train} illustrates how to train a token and sentence splitter with \texttt{TPipeline}.

\vspace{0.3cm}

\begin{figure*}[ht]
    \centering
    \includegraphics[scale=0.53]{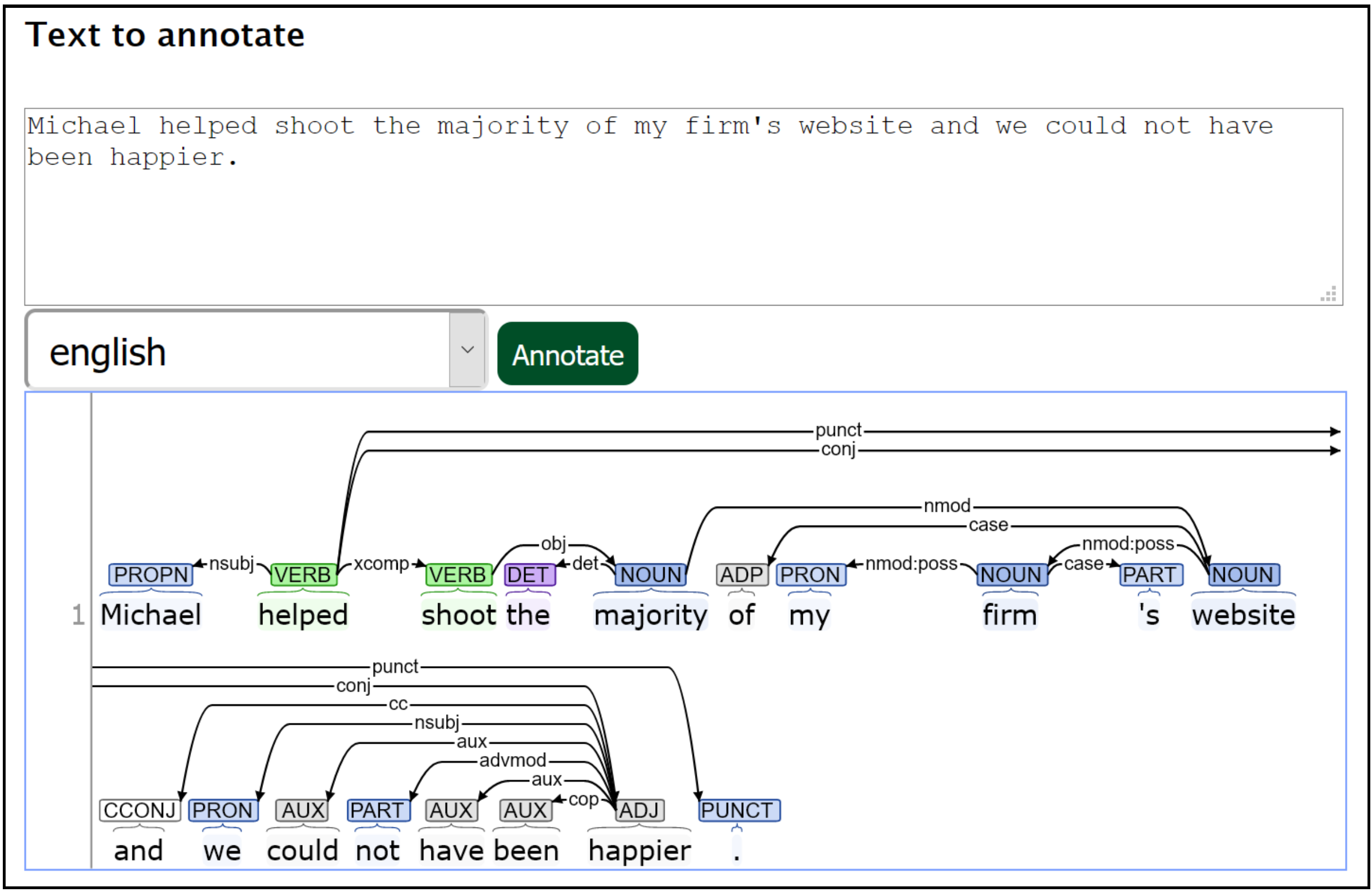}
    \caption{Demo website for Trankit.}
    \label{fig:demo-site}
\end{figure*}

\noindent \textbf{Demo Website}. A demo website for Trankit to support 90 pretrained pipelines is hosted at: \url{http://nlp.uoregon.edu/trankit}. Figure \ref{fig:demo-site} shows its interface.

\section{System Evaluation}
\label{sec:exp}

\subsection{Datasets \& Hyper-parameters}

To achieve a fair comparison, we follow Stanza \cite{qi-etal-2020-stanza} to train and evaluate all the models on the same canonical data splits of 90 Universal Dependencies treebanks v2.5 (UD2.5)\footnote{We skip 10 treebanks whose languages are not supported by \textit{XLM-Roberta}.} \citep{11234/1-3105}, and 11 public NER datasets provided in the following corpora: AQMAR \citep{mohit-etal-2012-recall}, CoNLL02 \citep{tjong-kim-sang-2002-introduction}, CoNLL03 \citep{tjong-kim-sang-de-meulder-2003-introduction}, GermEval14 \citep{benikova-etal-2014-nosta}, OntoNotes \citep{weischedel2013ontonotes}, and WikiNER \citep{nothman2012:artint:wikiner}. Hyper-parameters for all models and datasets are selected based on the development data in this work.

\begin{table*}[ht]
\centering
\addtolength{\belowcaptionskip}{-3mm}
\scalebox{0.8}{
\begin{tabular}{l|c|c|c|c|c|c|c|c|c}
\multicolumn{1}{c|}{System} & Tokens & Sents. & Words & UPOS  & XPOS  & UFeats & Lemmas & UAS   & LAS   \\ \hline
{\bf Trankit} (plug-and-play with adapters)                  & {\bf 99.05}  & {\bf 95.12}  & {\bf 98.96} & {\bf 95.43} & {\bf 89.02} & {\bf 92.69}  & {\bf 93.46}  & {\bf 86.20}  & {\bf 82.51} \\ \hline
Multilingual                   & 96.69  & 88.95  & 96.35 & 91.19 & 84.64 & 88.10  & 90.02  & 72.96 & 68.66 \\ \hline
No-adapters                          & 95.06  & 89.57  & 94.08 & 88.79 & 82.54 & 83.76  & 88.33  & 66.63 & 63.11 \\ 
\end{tabular}}
\caption{Model performance on 9 different treebanks (macro-averaged F1 score over test sets).}
\label{tab:ablation}
\end{table*}

\subsection{Universal Dependencies performance}

Table \ref{tab:ud-performance} compares the performance of Trankit and the latest available versions of other popular toolkits, including Stanza (v1.1.1) with current state-of-the-art performance, UDPipe (v1.2), and spaCy (v2.3) on the UD2.5 test sets.  The performance for all systems is obtained using the official scorer of the CoNLL 2018 Shared Task\footnote{\url{https://universaldependencies.org/conll18/evaluation.html}}. On five illustrated languages, Trankit achieves competitive performance on tokenization, MWT expansion, and lemmatization. Importantly, Trankit outperforms other toolkits over all remaining tasks (e.g., POS and morphological tagging) in which the improvement boost is substantial and significant for sentence segmentation and dependency parsing. For example, English enjoys a $7.22\%$ improvement for sentence segmentation, a $3.92\%$ and $4.37\%$ improvement for UAS and LAS in dependency parsing. For Arabic, Trankit has a remarkable improvement of $16.16\%$ for sentence segmentation while Chinese observes $12.31\%$ and $12.72\%$ improvement of UAS and LAS for dependency parsing.


Over all 90 treebanks, Trankit outperforms the previous state-of-the-art framework Stanza in most of the tasks, particularly for sentence segmentation ($+3.24\%$), POS tagging ($+1.44\%$ for UPOS and $+1.55\%$ for XPOS), morphological tagging ($+1.46\%$), and dependency parsing ($+4.0\%$ for UAS and $+5.01\%$ for LAS) while maintaining the competitive performance on tokenization, multi-word expansion, and lemmatization.

\subsection{NER results}

Table \ref{tab:ner-performance} compares Trankit with Stanza (v1.1.1), Flair (v0.7), and spaCy (v2.3) on the test sets of 11 considered NER datasets. Following Stanza, we report the performance for other toolkits with their pretrained models on the canonical data splits if they are available. Otherwise, their best configurations are used to train the models on the same data splits (inherited from Stanza). Also, for the Dutch datasets, we retrain the models in Flair as those models (for Dutch) have been updated in version v0.7. As can be seen, Trankit obtains competitive or better performance for most of the languages, clearly demonstrating the benefit of using the pretrained transformer for multilingual NER.



\begin{table}[ht]
\centering
\scalebox{0.75}{
\begin{tabular}{l|l|c|c|c|c}
Language                 & Corpus     & \multicolumn{1}{l|}{{\bf Trankit}} & \multicolumn{1}{l|}{Stanza} & \multicolumn{1}{l|}{Flair} & \multicolumn{1}{l}{spaCy} \\ \hline
Arabic                   & AQMAR      & \textbf{74.8}                & 74.3                        & 74.0                       & -                          \\ \hline
Chinese                  & OntoNotes  & \textbf{80.0}                & 79.2                        & -                          & 69.3                       \\ \hline
\multirow{2}{*}{Dutch}   & CoNLL02    & \textbf{91.8}                         & 89.2                        & 91.3                           & 73.8                       \\ \cline{2-6} 
                         & WikiNER    & \textbf{94.8}                         & \textbf{94.8}                        & \textbf{94.8}                        & 90.9                       \\ \hline
\multirow{2}{*}{English} & CoNLL03    & 92.1                & 92.1               & \textbf{92.7}                       & 81.0                       \\ \cline{2-6} 
                         & OntoNotes  & \textbf{89.6}                & 88.8                        & 89.0                       & 85.4                      \\ \hline
French                   & WikiNER    & 92.3                         & \textbf{92.9}               & 92.5                       & 88.8                      \\ \hline
\multirow{2}{*}{German}  & CoNLL03    & \textbf{84.6}                & 81.9                        & 82.5                       & 63.9                       \\ \cline{2-6} 
                         & GermEval14 & \textbf{86.9}                & 85.2                        & 85.4                       & 68.4                       \\ \hline
Russian                  & WikiNER    & 92.8                         & \textbf{92.9}               & -                          & -                          \\ \hline
Spanish                  & CoNLL02    & \textbf{88.9}                & 88.1                        & 87.3                       & 77.5                       \\ 
\end{tabular}}
\caption{Performance (F1) on NER test sets.}
\label{tab:ner-performance}
\end{table}

\begin{table}[ht]
\small 
\centering
\scalebox{1}{
\begin{tabular}{l|c|c|c|c}
\multirow{2}{*}{System} & \multicolumn{2}{c|}{GPU} & \multicolumn{2}{c}{CPU} \\ \cline{2-5} 
                        & UD          & NER        & UD          & NER        \\ \hline
Trankit                 & $4.50\times$       & $1.36\times$      & $19.8\times$       & $31.5\times$      \\ \hline
Stanza                  & $3.22\times$       & $1.08\times$      & $10.3\times$       & $17.7\times$      \\ \hline
UDPipe                  & -           & -          & $4.30\times$       & -          \\ \hline
Flair                   & -           & $1.17\times$      & -           & $51.8\times$      \\ 
\end{tabular}}
\caption{Run time on processing the English EWT treebank and the English Ontonotes NER dataset. Measurements are done on an NVIDIA Titan RTX card.}
\label{tab:speed}
\end{table}

\begin{table}[ht]
\centering
\small 
\begin{tabular}{l|c|c}
\multicolumn{1}{c|}{Model Package} & Trankit  & Stanza   \\ \hline
Multilingual Transformer                & 1146.9MB & -        \\ \hline
Arabic                              & 38.6MB   & 393.9MB  \\ \hline
Chinese                             & 40.6MB   & 225.2MB  \\ \hline
English                             & 47.9MB   & 383.5MB  \\ \hline
French                              & 39.6MB   & 561.9MB  \\ \hline
Spanish                             & 37.3MB   & 556.1MB  \\ \hline \hline
{\bf Total size}                          & 1350.9MB & 2120.6MB \\ 
\end{tabular}
\caption{Model sizes for five languages.}
\label{tab:model-size}
\end{table}

\subsection{Speed and Memory Usage}

Table \ref{tab:speed} reports the relative processing time for UD and NER of the toolkits compared to spaCy's CPU processing time\footnote{spaCy can process 8140 tokens and 5912 tokens per second for UD and NER, respectively.}. For memory usage comparison, we show the model sizes of Trankit and Stanza for several languages in Table \ref{tab:model-size}. As can be seen, besides the multilingual transformer, model packages in Trankit only take dozens of megabytes while Stanza consumes hundreds of megabytes for each package. This leads to the Stanza's usage of much more memory when the pipelines for these languages are loaded at the same time. In fact, Trankit only takes 4.9GB to load all the 90 pretrained pipelines for the 56 supported languages.



\subsection{Ablation Study}

This section compares Trankit with two other possible strategies to build a multilingual system for fundamental NLP tasks. In the first strategy (called ``\textit{Multilingual}''), we train a single pipeline where all the components in the pipeline are trained with the combined training data of all the languages. The second strategy (called ``\textit{No-adapters}'') involves eliminating adapters from \textit{XLM-Roberta} in Trankit. As such, in ``\textit{No-adapters}'', pipelines are still trained separately for each language; the pretrained transformer is fixed; and only task-specific weights (for predictions) in components are updated during training.


%

For evaluation, we select 9 treebanks for 3 different groups, i.e., high-resource, medium-resource, and low-resource, depending on the sizes of the treebanks. In particular, the high-resource group includes Czech, Russian, and Arabic; the medium-resource group includes French, English, and Chinese; and the low-resource group involves Belarusian, Telugu, and Lithuanian. Table \ref{tab:ablation} compares the average performance of Trankit, ``\textit{Multilingual}'', and  ``\textit{No-adapters}''. As can be seen, ``\textit{Multilingual}'' and  ``\textit{No-adapters}'' are significantly worse than the proposed adapter-based Trankit. We attribute this to the fact that multilingual training might suffer from unbalanced sizes of treebanks, causing high-resource languages to dominate others and impairing the overall performance. For ``\textit{No-adapters}'', fixing pretrained transformer might significantly limit the models' capacity for multiple tasks and languages.





\section{Conclusion and Future Work}
We introduce Trankit, a transformer-based multilingual toolkit that significantly improves the performance for fundamental NLP tasks, including sentence segmentation, part-of-speech, morphological tagging, and dependency parsing over 90 Universal Dependencies v2.5 treebanks of 56 different languages. Our toolkit is fast on GPUs and efficient in memory use, making it usable for general users. In the future, we plan to improve our toolkit by investigating different pretrained transformers such as mBERT and $\textrm{XLM-Roberta}_{large}$. We also plan to provide Named Entity Recognizers for more languages and add modules to perform more NLP tasks.


\section*{Acknowledgments}

This research has been supported by the Office of the Director of National Intelligence (ODNI), Intelligence Advanced Research Projects Activity (IARPA), via IARPA Contract No. 2019-19051600006 under the Better Extraction from Text Towards Enhanced Retrieval (BETTER) Program. The views and conclusions contained herein are those of the authors and should not be interpreted as necessarily representing the official policies, either expressed or implied, of ODNI, IARPA, the Department of Defense, or the U.S. Government. The U.S. Government is authorized to reproduce and distribute reprints for governmental purposes notwithstanding any copyright annotation therein. This document does not contain technology or technical data controlled under either the U.S. International Traffic in Arms Regulations or the U.S. Export Administration Regulations.

\bibliography{anthology,eacl2021}
\bibliographystyle{acl_natbib}

\end{document}